%% file: main.tex
\documentclass[sigconf]{acmart}

\AtBeginDocument{%
  }

\copyrightyear{2026}
\acmYear{2026}
\setcopyright{cc}
\setcctype{by}
\acmConference[KDD '26]{Proceedings of the 32nd ACM SIGKDD Conference on Knowledge Discovery and Data Mining V.2}{August 09--13, 2026}{Jeju Island, Republic of Korea}
\acmBooktitle{Proceedings of the 32nd ACM SIGKDD Conference on Knowledge Discovery and Data Mining V.2 (KDD '26), August 09--13, 2026, Jeju Island, Republic of Korea}
\acmDOI{10.1145/3770855.3818311}
\acmISBN{979-8-4007-2259-2/2026/08}

\usepackage{graphicx}
\usepackage{multirow}
\usepackage{balance}

\begin{document}

%%
%% The "title" command has an optional parameter,
%% allowing the author to define a "short title" to be used in page headers.
\title{A Foundation Model for Multimodal Event Sequences in Financial Applications}

%%
%% The "author" command and its associated commands are used to define
%% the authors and their affiliations.
%% Of note is the shared affiliation of the first two authors, and the
%% "authornote" and "authornotemark" commands
%% used to denote shared contribution to the research.

\author{Nikita Rusakov}
\orcid{0009-0000-3145-4724}
\affiliation{
  \institution{Sber}
  \city{Moscow}
  \country{Russian Federation}
}
\email{nanrusakov@gmail.com}
\authornote{Authors contributed equally to the paper}

\author{Vladislav Meshkov}
\orcid{0009-0003-1393-0522}
\affiliation{
  \institution{Sber}
  \city{Moscow}
  \country{Russian Federation}
}
\email{vladmeshkov160@gmail.com}
\authornotemark[1]

\author{Konstantin Zorin}
\orcid{0009-0001-3408-4199}
\affiliation{
  \institution{Sber}
  \city{Moscow}
  \country{Russian Federation}
}
\email{kostya.zorin.2003.ri@gmail.com}
\authornotemark[1]

\author{Gleb Zaripov}
\orcid{0009-0009-0388-1417}
\affiliation{
  \institution{Sber}
  \city{Moscow}
  \country{Russian Federation}
}
\email{gleb.zar.030@gmail.com}
\authornotemark[1]

\author{Alexander Uglov}
\affiliation{
  \institution{Sber}
  \city{Moscow}
  \country{Russian Federation}
}
\email{skilletthebest25@gmail.com}
\authornotemark[1]

\author{Alexey Vasilev}
\orcid{0009-0007-1415-2004}
\affiliation{
  \institution{Sber AI Lab}
  \city{Moscow}
  \country{Russian Federation}
}
\email{alexxl.vasilev@yandex.ru}

\author{Anton Klenitskiy}
\orcid{0009-0005-8961-6921}
\affiliation{
  \institution{Sber AI Lab}
  \city{Moscow}
  \country{Russian Federation}
}
\email{antklen@gmail.com}

%%
%% By default, the full list of authors will be used in the page
%% headers. Often, this list is too long, and will overlap
%% other information printed in the page headers. This command allows
%% the author to define a more concise list
%% of authors' names for this purpose.
\renewcommand{\shortauthors}{Nikita Rusakov et al.}

%%
%% The abstract is a short summary of the work to be presented in the
%% article.
\begin{abstract}

\input{sections/0_abstract}
\end{abstract}

%%
%% The code below is generated by the tool at http://dl.acm.org/ccs.cfm.
%% Please copy and paste the code instead of the example below.
%%


%%
%% Keywords. The author(s) should pick words that accurately describe
%% the work being presented. Separate the keywords with commas.
%% A "teaser" image appears between the author and affiliation
%% information and the body of the document, and typically spans the
%% page.
% \begin{teaserfigure}
%   \includegraphics[width=\textwidth]{sampleteaser}
%   \caption{Seattle Mariners at Spring Training, 2010.}
%   \Description{Enjoying the baseball game from the third-base
%   seats. Ichiro Suzuki preparing to bat.}
%   \label{fig:teaser}
% \end{teaserfigure}

% \received{20 February 2007}
% \received[revised]{12 March 2009}
% \received[accepted]{5 June 2009}

%%
%% This command processes the author and affiliation and title
%% information and builds the first part of the formatted document.
\maketitle

\section{Introduction}

\input{sections/1_intro}

\section{Related Work}

\input{sections/2_related_work}

\section{Problem Formulation}

\input{sections/3_problem_formulation}

\section{Approach}

\input{sections/4_0_approach}
\input{sections/4_sequential_model}

\input{sections/5_tabular_model}

\section{Offline Experiments}

\input{sections/6_offline_experiments}

\section{Online Results}

\input{sections/7_online_results}

\section{Conclusion}

\input{sections/8_conclusion}

%%
%% The next two lines define the bibliography style to be used, and
%% the bibliography file.
\bibliographystyle{ACM-Reference-Format}
\balance
\bibliography{sections/bibliography}

\end{document}

%% file: sections/0_abstract.tex
Predictive modeling is a core component of modern financial services, where a wide range of tasks are traditionally addressed using separate models trained on manually engineered tabular features. This task-specific approach limits reuse and makes it difficult to fully exploit heterogeneous data sources such as transaction histories and digital interaction signals. In this paper, we present an approach based on pretraining a foundation transformer model on multimodal sequences of user events. Events from multiple data sources are unified into a single chronological sequence, enabling early fusion of heterogeneous modalities and learning of general-purpose representations via a next-event prediction objective. These representations are combined with existing engineered user features, on top of which lightweight neural models are trained for multiple downstream tasks. The proposed system outperforms traditional task-specific models while reducing development overhead. The approach was deployed in production at one of the biggest banks in Eastern Europe, resulting in measurable improvements in business metrics.

%% file: sections/1_intro.tex
Machine learning is a core component of modern financial services, supporting a wide range of predictive tasks such as risk assessment, fraud detection, product recommendations, and customer analytics~\cite{braithwaite2025your,yeh2025treasure,skalski2023towards,babaev2022coles,jha2012employing}. In large financial institutions, these tasks are typically addressed by building separate models for each use case, trained on manually engineered features derived from user transaction histories and digital interaction data. While this task-specific approach has demonstrated strong performance in individual applications, it leads to significant duplication of effort, limited reuse of learned representations, and increasing system complexity as the number of predictive tasks grows.

A key limitation of this paradigm is its inability to fully leverage the scale and richness of available data~\cite{braithwaite2025your,yeh2025treasure}. In practice, user behavior is recorded across multiple heterogeneous sources, including financial transactions, online interactions, communications, and other behavioral signals, each characterized by different structures and temporal dynamics~\cite{mollaev2025multimodal}. Large financial institutions collect massive volumes of such event-level data over long time horizons. Transforming these multimodal event streams into task-specific tabular features requires substantial manual effort and often leads to information loss. As a result, only a fraction of the available data is effectively utilized, and insights learned for one task are difficult to transfer or reuse across others.

To address these challenges, we explore an approach based on pretraining of a transformer-based foundation model on sequences of user events~\cite{braithwaite2025your,yeh2025treasure,karpukhin2025ht,ostroukhov2026pragma,dou2025transactiongpt}. The model is trained in a self-supervised manner to learn general-purpose representations from large-scale heterogeneous data that can be reused across multiple downstream tasks. Events from different data sources are unified into a single chronological sequence, enabling early fusion of multimodal signals, and the model is pretrained using a next-event prediction objective. The pretrained model serves as a shared representation backbone: its embeddings are frozen and combined with existing engineered user features, on top of which lightweight task-specific tabular neural models are trained for individual applications.

We evaluate the proposed approach across multiple real-world financial prediction tasks at Sber, one of the biggest banks in Eastern Europe, serving more than 100 million customers. In our experiments, we demonstrate that self-supervised pretraining and the use of multiple data modalities consistently improve predictive performance compared to traditional task-specific models. We further analyze the impact of different fusion strategies and study scaling behavior with respect to model size and sequence length under industrial inference constraints. The proposed system has been deployed in production, resulting in measurable improvements in business metrics while reducing model development time and overall system complexity.

%% file: sections/2_related_work.tex
\subsection{Foundation models for user behavior modeling and recommendation systems}

Several recent industrial works explore pretrained transformer-based models for modeling user behavior. These approaches focus on learning reusable user representations from sequences of user actions, which can be applied across multiple downstream tasks. Pinnerformer~\cite{pancha2022pinnerformer} introduces a transformer-based model trained on user actions to produce high-quality user embeddings that can be used for many recommendation tasks at Pinterest. TransAct~\cite{xia2023transact} proposes a hybrid ranking framework that combines short-term user interests captured by a transformer-based real-time sequential model with long-term user representations computed offline using PinnerFormer. Finally, PinFM~\cite{chen2025pinfm} explicitly adopts the foundation model paradigm and pretraining-finetuning approach for user activity modeling at Pinterest. The authors pretrain a sequential model on massive user activity sequence data collected across multiple applications. Downstream ranking models reuse the pretrained transformer and embedding tables, which are further fine-tuned with application-specific data and features. ARGUS~\cite{khrylchenko2025scaling} studies the scalability of autoregressively pretrained transformer models for recommendations. The model is pretrained on user histories using next-item and feedback prediction objectives and then fine-tuned for various ranking tasks. The work demonstrates that this approach scales effectively across a wide range of transformer sizes. DV365~\cite{lyu2025dv365} focuses on encoding extremely long user histories into compact user embeddings. The proposed offline foundational model precomputes user representations from long behavior sequences, which are then shared across a large number of downstream models, improving performance across corresponding tasks.

Universal behavioral modeling was also the focus of the RecSys Challenge 2025~\cite{dabrowski2025recsys}. The goal was to learn task-independent user representations from rich multi-event user logs such as purchases, cart additions, page visits, and search queries, which are applicable across different downstream tasks, including churn prediction, product propensity, and category propensity prediction. The work~\cite{sawada2025toward} proposes a contrastive learning approach with a transformer model pretrained to maximize agreement between representations of past and future segments of the same user’s behavior, while pushing apart representations of different users. Another solution~\cite{makeev2025blending} is based on pretraining a large transformer model using the next-event prediction objective, where several attributes of the next event are predicted jointly. The paper~\cite{klenitskiy2025encode} proposes learning universal representations using a sequential autoencoder that reconstructs the whole user history from the user embedding. Beyond the RecSys Challenge, several other works also explore pretraining approaches for user behavior modeling~\cite{wang2023sequence,gong2025behavegpt,liu2022user}.

\subsection{Representation learning from transaction sequences for financial applications}

Autoregressive pretraining and contrastive learning are among the most widely used approaches for representation learning on transactional data~\cite{bazarova2025learning,sakhno2025pytorch,moskvoretskii2024mlem}.
Early work on learning general-purpose embeddings from user transaction histories primarily relied on recurrent neural networks. CoLES~\cite{babaev2022coles} represents an example of the contrastive learning paradigm for event sequence modeling. The method learns embeddings for subsequences of user event histories by encouraging representations of subsequences from the same user to be similar, while pushing apart representations of different users. Embeddings learned from transaction datasets are shown to be effective for downstream financial tasks. NPPR~\cite{skalski2023towards} follows an autoregressive pretraining approach, training an RNN-based model on large corpora of transaction sequences using a combination of next-transaction prediction and past reconstruction objectives.  The authors demonstrate the success of the next-event prediction task for financial applications.

More recent works apply transformer-based autoregressive pretraining to large-scale transactional datasets~\cite{yeh2025treasure,braithwaite2025your,dou2025transactiongpt}. For example, TREASURE~\cite{yeh2025treasure} introduces a transformer foundation model specifically designed for transaction data. The model is pretrained using next-transaction prediction along with an auxiliary objective capturing current transaction network signals. In nuFormer~\cite{braithwaite2025your}, the authors investigate transformer-based representation learning for transaction data with both textual and structured attributes. The authors propose a tokenization scheme where each transaction is represented by multiple tokens and pretrain causal transformers using the standard next-token prediction objective. An end-to-end fine-tuning approach is introduced to combine learned user embeddings with existing tabular features for downstream financial tasks. Another line of work explores the use of large language models for transactional data~\cite{polleti2025open,guo2025efficient,raman2024scalable,shestov2025llm4es}.

In summary, prior work has demonstrated the effectiveness of pretrained sequential models for user behavioral modeling and transaction-based financial applications. Existing approaches either focus on non-financial user interaction data or model transaction sequences in isolation. In contrast, our work addresses the practical setting of large-scale financial systems, where the data is inherently multimodal, since user behavior is observed across multiple heterogeneous data sources.

%% file: sections/3_problem_formulation.tex
We consider the problem of learning general-purpose user representations from large-scale, heterogeneous behavioral data in a financial system. User behavior is recorded across multiple data sources, each capturing complementary aspects of user activity. For a given user \( u \), each modality \( m\) from a set of available modalities \( M \)  produces a temporal sequence of events
\[ S_u^{(m)} = [e_{u1}^{(m)}, e_{u2}^{(m)}, \dots]. \]
Events from different modalities are heterogeneous in structure and attributes, but share a common timeline. A key challenge is to integrate these heterogeneous sequences into a unified representation. Existing approaches typically rely on either late fusion, in which each modality is modeled independently, or early fusion, which merges events from all modalities into a single chronological sequence. While late fusion limits cross-modal temporal interactions, early fusion requires a unified event representation that can handle heterogeneous inputs.

In addition to raw event sequences, production systems rely on a rich set of engineered user features  \(\mathbf{f}_u \in R^K\). These features encode domain knowledge, have strong predictive value in existing tabular models, and
are already deeply integrated into downstream pipelines. A practical solution should therefore complement, rather than replace them.

The learned representations should apply to diverse downstream business tasks without task-specific adaptation. In practical settings, this task set is not fixed in advance, is typically large, and evolves over time, since new tasks may emerge as business objectives change. Training and maintaining a separate sequence model for each task is therefore computationally expensive and operationally impractical. To summarize, our objective is to learn a universal user embedding function
\[
\Phi_{\text{core}}: \{S_u^{(1)}, \dots, S_u^{(M)} \} \rightarrow \mathbf{z}_u \in \mathbb{R}^D,
\]
such that:
\begin{itemize}
    \item \( \mathbf{z}_u \) captures behavioral patterns and temporal dynamics across all modalities;
    \item the embedding is task-independent and can be reused across diverse downstream objectives;
    \item  \( \Phi_{\text{core}} \) can be frozen after pretraining, enabling lightweight downstream models to be trained efficiently on top of \( (\mathbf{z}_u, \mathbf{f}_u) \) without fine-tuning the core model.
\end{itemize}

%% file: sections/4_0_approach.tex
We adopt an early fusion strategy to capture cross-modal interactions in user behavior better. Events from all available data sources are merged into a single chronological sequence, and we pretrain a transformer-based sequence encoder on top of it using the next-event prediction objective. This pretrained encoder serves as a shared representation backbone: its parameters are frozen and reused without fine-tuning across all downstream tasks. For downstream applications, the sequence embedding \( \mathbf{z}_u \) is combined with existing engineered features \( \mathbf{f}_u \). 
Features \( \mathbf{f}_u \) are processed by a separate transformer-based tabular model, whose output embedding is concatenated with the sequence embedding. A lightweight MLP head is added to predict the final objective. The following sections describe the components of this pipeline in detail.

%% file: sections/4_sequential_model.tex
\subsection{Pretrained Sequence Encoder}

\begin{figure*}[t]
    \centering
    \includegraphics[width=\textwidth]{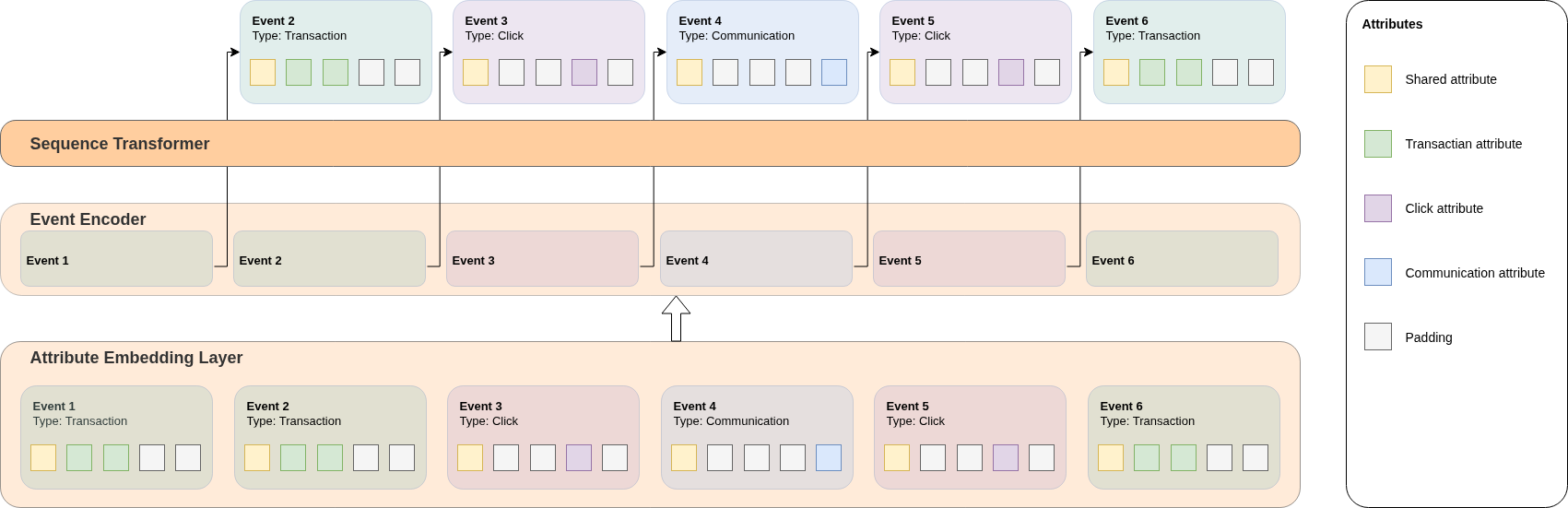}%
    \caption{Pretrained sequence model architecture. The sequential processing of heterogeneous user events (transactions, clicks, communications) via an event encoder and a shared Transformer backbone, followed by event-specific predictive heads that jointly model multiple target attributes for each next-event type.
    }
    \Description{Block diagram of the pretrained sequence model, with data flowing from bottom to top. At the input there is a chronological sequence of heterogeneous user events of different types: transactions, clicks, and communications. Each event first passes through a shared event encoder that maps the different event types into a common embedding space. The resulting sequence of event embeddings is fed into a  Transformer backbone, which processes the events jointly and produces a representation for each position. On top of the backbone, multiple event-specific predictive heads sit, one per next-event type. Each head jointly predicts several target attributes that characterize the corresponding next event.}
    \label{fig:next_event_model}
\end{figure*}
% Here, we denote transactions and communications as TXN and Comm, respectively.

We use a three-stage architecture to process heterogeneous, multimodal event sequence data. The architecture consists of an \textit{Embedding Layer}, an \textit{Event Encoder}, and a \textit{Transformer Backbone}, as shown in Figure \ref{fig:next_event_model}.

\subsubsection{Embedding Layer}

Events are represented by a set of numerical and categorical attributes. The embedding layer maps individual event attributes into a common embedding dimension, producing fixed-size representations that are later combined at the event level.

For numerical attributes, we first apply z-score normalization:
\[
\tilde{x}_{\text{num}} = \frac{x_{\text{num}} - \mu_{\text{num}}}{\sigma_{\text{num}}},
\]
where \(x_{\text{num}} \in \mathbb{R}\) denotes a raw numerical value, and $\mu_{\text{num}}$ and $\sigma_{\text{num}}$ are the corresponding mean and standard deviation. The normalized value is then projected into a $d$-dimensional embedding space using a learnable linear transformation:
\[
\mathbf{a}_{\text{num}} = \mathbf{w}_{\text{num}} \cdot \tilde{x}_{\text{num}} + \mathbf{b}_{\text{num}},
\]
where \(\mathbf{w}_{\text{num}} \in \mathbb{R}^{d}\) and \(\mathbf{b}_{\text{num}} \in \mathbb{R}^{d}\) are learnable parameters.

Categorical attributes are mapped using learnable embedding tables with the same embedding dimension $d$.

\subsubsection{Event Encoder: heterogeneous attribute fusion}

% Moreover,  events of different types often have diverse numbers of attributes, which the layer must handle coherently. 

A key challenge in modeling heterogeneous event data is that different event types are characterized by different sets of attributes. For example, a transaction event may include amount, location, and category attributes, whereas a clickstream event may contain application section, device type, and related metadata. The proposed Event Encoder addresses this heterogeneity by padding, modeling intra-event interactions between attributes, and aggregating them into a single fixed-size event representation.

\paragraph{Padding}
Each event \(e\) is represented by a variable-length set of attributes, where the number of attributes depends on the modality. After embedding each attribute into \(\mathbb{R}^{d}\), the set is padded to a fixed attribute dimension \(N_{\text{attr}}\), allowing events of different types to be stacked into a tensor of fixed shape for batch training. Formally, each event is represented as a matrix
\[
\mathbf{A}_e = [\mathbf{a}_1, \mathbf{a}_2, \ldots, \mathbf{a}_{N_{\text{attr}}}] \in \mathbb{R}^{N_{\text{attr}} \times d},
\]
where \(\mathbf{a}_i \in \mathbb{R}^{d}\) is the embedding of \(i\)-th attribute, and padded positions correspond to missing attributes and are masked during subsequent computations. Here, \(N_{\text{attr}}\) denotes the total number of unique attributes among all events used to form a representation.

\paragraph{Intra-event self-attention across attributes.}
To capture dependencies between attributes and to identify the most informative ones, we apply self-attention along the attribute dimension. This mechanism is conceptually related to intra-feature attention used in TP-BERTa \cite{yan2024makingpretrainedlanguagemodels}. Given the attribute matrix \(\mathbf{A}_e\), we compute
\[
\mathbf{A}'_e = \mathrm{Attn}(\mathbf{A}_e)
=\mathrm{Softmax}\!\left(\frac{\mathbf{Q}\mathbf{K}^{\top}}{\sqrt{d}}\right)\mathbf{V},
\]
where \(\mathbf{Q},\mathbf{K},\mathbf{V}\in\mathbb{R}^{N_{\text{attr}} \times d}\) are learned linear projections of \(\mathbf{A}_e\). Attention scores corresponding to padded positions are masked before the softmax operation so that missing attributes do not influence the resulting representations.

% Let \(\mathbf{A}'_e=\mathrm{Attn}(\mathbf{A}_e)\in\mathbb{R}^{L_{\max}\times d}\), and denote its \(i\)-th row (the refined attribute vector).

\paragraph{Event aggregation}
The refined attribute representations are aggregated into a single event-level embedding by pooling over the attribute dimension, followed by LayerNorm:
\[
\mathbf{e}
=\mathrm{LayerNorm}\!\left(\frac{1}{\sum_{i=1}^{N_{\text{attr}}} m_i}\sum_{i=1}^{N_{\text{attr}}} m_i\,\mathbf{a}'_i\right),
\]
where \(\mathbf{a}'_i\in\mathbb{R}^{d}\) denotes the attribute representation after self-attention ($i$-th row of \(\mathbf{A}'_e\)), and \(m_i\in\{0,1\}\) is a binary mask indicating non-padding positions. The normalization ensures that padding does not affect the aggregated representation.

Finally, we concatenate event embeddings to construct the event sequence:
\[
\mathbf{X} = [\mathbf{e}_1, \mathbf{e}_2, \ldots, \mathbf{e}_T] \in \mathbb{R}^{T \times d},
\]
where \(T\) is the number of events in the sequence. This yields a homogeneous sequence representation despite heterogeneous event attribute structure.

\subsubsection{Transformer Backbone: sequence modeling}

The sequence of event embeddings is processed by a causal transformer backbone that closely follows the GPT-2 architecture~\cite{radford2019language}. It consists of masked multi-head self-attention and position-wise feedforward layers with residual connections and layer normalization, enabling autoregressive modeling of event sequences.

The output of the final transformer layer produces contextualized representations for each event in the sequence. The final user representation \( \mathbf{z}_{\text{seq}}\) is obtained by mean pooling over the temporal dimension of these outputs.
 
\subsubsection{Pretraining objective}  

The model is trained using a \textit{next-event prediction} objective. Given the hidden state (the output of the last transformer layer) \(\mathbf{h}_t \in \mathbb{R}^{d}\) corresponding to the event at position \(t\), the model predicts the attributes of the subsequent event at position \(t+1\).

For each categorical attribute \(c\) of the next event, we apply a linear projection followed by a softmax to obtain a vector of predicted probabilities for all categories:
\[
\hat{\mathbf{p}}^{(c)}_{t+1} = \mathrm{Softmax}\!\left( W^{(c)}_{\mathrm{cat}} \mathbf{h}_t + \mathbf{b}^{(c)}_{\mathrm{cat}} \right),
\]
where \(W^{(c)}_{\mathrm{cat}} \in \mathbb{R}^{C_c \times d}\), \(\mathbf{b}^{(c)}_{\mathrm{cat}} \in \mathbb{R}^{C_c}\), and \(C_c\) denotes the number of categories for the categorical attribute \(c\).

For each numerical attribute \(n\), we use a linear projection to obtain a scalar prediction:
\[
\hat{y}^{(n)}_{t+1} = \mathbf{w}^{(n)}_{\text{num}} \mathbf{h}_t + b^{(n)}_{\text{num}},
\]
where \(\mathbf{w}^{(n)}_{\text{num}} \in \mathbb{R}^{d}\) and \(b^{(n)}_{\text{num}} \in \mathbb{R}\).

The training loss combines cross-entropy (CE) losses for categorical attributes and mean squared error (MSE) losses for numerical attributes. For each position \(t\), the loss is computed only over attributes actually present in the next event, ignoring missing attributes associated with other modalities. Let \(\mathcal{C}_{t+1}\) and \(\mathcal{N}_{t+1}\) denote the sets of observed categorical and numerical attributes of the event at \(t+1\). The loss at position \(t\) is defined as:
\[
\mathcal{L}_t =
\sum_{c \in \mathcal{C}_{t+1}} \text{CE}\!\left(\mathbf{y}^{(c)}_{t+1}, \hat{\mathbf{p}}^{(c)}_{t+1}\right)
+ \sum_{n \in \mathcal{N}_{t+1}} \text{MSE}\!\left(y^{(n)}_{t+1}, \hat{y}^{(n)}_{t+1}\right),
\]
where \(\mathbf{y}^{(c)}_{t+1}\) is the one-hot target for categorical attribute \(c\), and \(y^{(m)}_{t+1}\) is the target value for numerical attribute \(n\). The final objective aggregates \(\mathcal{L}_t\) over all valid positions in a sequence.

%% file: sections/5_tabular_model.tex
\subsection{Tabular Model}

\begin{figure}[htbp]
    \centering
    \includegraphics[scale=0.35]{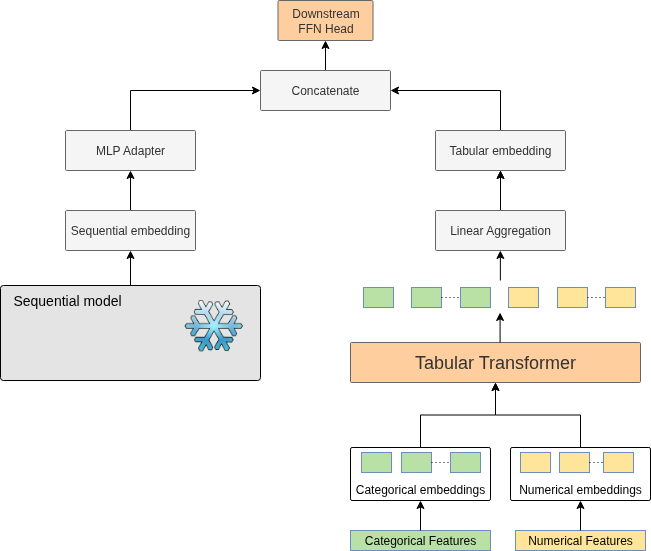}
    \caption{Schematic overview of the proposed downstream model architecture. The pipeline integrates a frozen sequential model with a tabular Transformer. Sequential events are encoded into embeddings and refined via an MLP adapter. Simultaneously, the tabular Transformer processes both categorical and numerical engineered features, and its output is aggregated. In the end, two representations are concatenated, and a task-specific head is added.}
    \Description{Block diagram of the downstream model architecture, showing two parallel branches that merge before the output. In the first branch, a frozen pretrained sequential model encodes user events into embeddings, which are then refined by an MLP adapter. In the second branch, a tabular Transformer processes engineered categorical and numerical features, and its output is aggregated into a single vector. The two resulting representations are concatenated, and a task-specific head is added on top to produce the prediction.}
    \label{fig:model_architecture}
\end{figure}

We apply a transformer-based neural network for tabular data, hereafter referred to as \textit{TabNN}, to process engineered user-level features and produce a tabular embedding for downstream prediction. The tabular representation is constructed from two types of input features: categorical and numerical. The model architecture is illustrated in Figure~\ref{fig:model_architecture}.

\subsubsection{Input preprocessing}

Numerical features undergo a two-stage normalization procedure. First, a signed logarithmic transform is applied to reduce skewness: \(x' = \text{sign}(x)\cdot\log(1 + |x|)\). The result is then standardized using the mean (\(\mu\)) and standard deviation (\(\sigma\)) calculated from the training data: \(\tilde{x} = (x' - \mu) / (\sigma + \varepsilon)\), where \(\varepsilon > 0\) is a small constant added for numerical stability.

\subsubsection{Embedding Layer}

Each numerical feature is associated with a trainable  vector \( \mathbf{w}_{\text{num}}^j \in \mathbb{R}^{d_{\text{emb}}} \), which is scaled by the corresponding normalized feature value:

\[
\mathbf{e}_{\text{num}}^j = \mathbf{w}_{\text{num}}^j \cdot \tilde{x}^j,
\]
where \( \tilde{x}^j \in \mathbb{R} \) denotes the normalized value of the $j$-th numerical feature. The total number of numerical features is denoted by \( N_{\text{\text{num}}} \).

For each categorical feature, a trainable embedding table is learned, assigning a vector of dimension \( d_{\text{emb}} \) to each category. Let us denote the embedding of the \( j \)-th categorical feature as \( \mathbf{e}_{\text{cat}}^j \), and the number of categorical features is $N_{\text{cat}}$.

\subsubsection{Concatenation \& Noise Injection}

All feature embeddings are concatenated into a single sequence:
\[ \mathbf{E}_{\text{concat}} = [\mathbf{e}^1_{\text{\text{cat}}}, \dots, \mathbf{e}^{N_{\text{cat}}}_{\text{cat}}, \mathbf{e}^1_{\text{num}}, \dots, \mathbf{e}^{N_{\text{num}}}_{\text{num}}].\]
To improve robustness and provide regularization, a small random noise vector is added to each embedding in the sequence during training.

\subsubsection{Transformer Processing}

The resulting sequence of feature embeddings is processed by a Transformer encoder. Each layer applies multi-head self-attention \textit{across the feature dimension}, allowing the model to learn interactions between different user features.

\subsubsection{Weighted Aggregation}

The sequence of feature representations \( \mathbf{H} = [\mathbf{h}_1, \mathbf{h}_2, \dots, \mathbf{h}_{N_{\text{total}}}] \) from the final Transformer layer is aggregated into a single fixed-size vector \( \mathbf{z}_{\text{tab}} \) using a weighted sum with learnable weights \( \boldsymbol{\alpha} \in \mathbb{R}^{N_{\text{total}}} \):
\[
\mathbf{z}_{\text{tab}}  = \sum_{i=1}^{N_{\text{total}}} \alpha_i \cdot \mathbf{h}_i,
\]
where \( N_{\text{total}} = N_{\text{num}} + N_{\text{cat}}\) is the total number of features.

\subsection{Downstream pipeline}

For downstream tasks, the final user representation is obtained by combining embeddings from the sequential and tabular models, as illustrated in Figure~\ref{fig:model_architecture}. The sequential user embedding is first adapted to the target dimensionality using a lightweight MLP adapter consisting of two linear layers with LayerNorm, SeLU activation\cite{klambauer2017selfnormalizingneuralnetworks}, and dropout:
\[
\mathbf{z}_{\text{seq}}' = \text{MLP}_{\text{adapter}}(\mathbf{z}_{\text{seq}}).
\]

The adapted sequential embedding \( \mathbf{z}_{\text{seq}}' \) is then concatenated with the tabular embedding \( \mathbf{z}_{\text{tab}} \) produced by TabNN:
\[
\mathbf{z}_u = [\mathbf{z}_{\text{seq}}', \mathbf{z}_{\text{tab}}].
\]

The combined representation \( \mathbf{z}_u \) is passed to a task-specific head (e.g., a linear layer or MLP) for final prediction:
\[
\hat{y} = \text{Head}_{\text{task}}(\mathbf{z}_u).
\]

To adapt the system to a particular downstream task, the pretrained sequential encoder is kept \textit{frozen}. Training is performed only on the parameters of the TabNN and the MLP adapter, as well as the task-specific head. This approach ensures that task-relevant patterns are extracted from the aggregate features, while the large, general-purpose sequential embedding provides a stable, information-rich foundation.

As a baseline for evaluating the contribution of the pretrained sequential model, we use the same TabNN architecture trained solely on engineered features with the corresponding task-specific head, without incorporating the sequential embedding \( \mathbf{z}_{\text{seq}}' \). In addition, for ablation analysis, we consider a complementary setting in which only the sequential embedding \( \mathbf{z}_{\text{seq}}' \) is used together with a task-specific head, excluding the tabular embedding \( \mathbf{z}_{\text{tab}} \).

\subsection{Design Choices}

\subsubsection{Pretraining objective}

The choice of the next-event prediction (NEP) objective for pretraining the sequential model is motivated by prior work demonstrating the effectiveness of next-step prediction for learning transferable representations from behavioral sequences~\cite{khrylchenko2025scaling,braithwaite2025your,skalski2023towards,yeh2025treasure}. In addition, we empirically evaluated several alternative self-supervised objectives, including CoLES~\cite{babaev2022coles}, masked language modeling (MLM), and DeTPP~\cite{karpukhin2024detecting}, and NEP yielded the strongest performance. 

\subsubsection{Tabular model architecture}

The described tabular processing pipeline was selected through extensive evaluation on our proprietary datasets. We benchmarked several state-of-the-art approaches, including standard Feed-Forward Neural Networks, Deep \& Cross Network v2 (DCNv2) \cite{inproceedings_dcn}, and the recently proposed TabM model \cite{gorishniy2025tabm}. Our architecture is closely related to the FT-Transformer \cite{ft_transformer} but differs in the choice of activation functions and the order of normalization operations. In our experiments, the selected TabNN architecture outperformed other approaches across multiple tasks.

%% file: sections/6_offline_experiments.tex
We evaluate the proposed approach through a series of offline experiments, including comparison with a strong feature-based baseline, analysis of scaling behavior, assessment of self-supervised pretraining, and ablations on input modalities and fusion design.

\subsection{Experimental Setup}

\subsubsection{Datasets}

We use three behavioral data domains to construct multimodal sequential inputs for foundation model pretraining:

\begin{itemize}
\item Transactions (Txn): financial operations (debits, credits, transfers) represented by attributes including transaction amount, category, and timestamp.
\item Clickstream (Click): time-ordered sequences of user interactions within the bank's mobile application, including visited application sections and associated timestamps.
\item Communications (Comm): records of customer interactions with the bank, including chat and call-center requests, push notifications, emails, and product applications.
\end{itemize}

Their feature composition and sequence statistics are summarized in Table ~\ref{tab:stats}. All datasets cover 2 years.

\begin{table}[ht]
\centering
\caption{Feature set statistics by data domain. Seq. length/year denotes the length of a customer event sequence per year.}

\label{tab:stats}
\begin{tabular}{lccc}
\hline
\textbf{Statistic} & \textbf{Txn} & \textbf{Click} & \textbf{Comm} \\
\hline
Mean seq. length/year            & 717   & 273  & 115 \\
Median seq. length/year          & 489   & 43   & 87  \\
\# Numerical features      & 1     & 0    & 0   \\
\# Categorical features    & 7     & 1    & 5   \\
\# Unique category values  & 945 & 230 & 159 \\
Events (billions)          & 160   & 30   & 20  \\
\hline
\end{tabular}
\end{table}

% \begin{tabular}{llccc}
% \hline
% && Txn & Clickstream & Comm \\
% \hline
%  \multirow{2}{*}{Sequence length}& Mean  & 717 & 273 & 115 \\
% per year &Median  & 489 & 43 & 87 \\
% \multirow{3}{*}{Count features}&  num features & 1 & 0 & 0 \\
% &\cat features & 7 & 1 & 5 \\
% &\# cat values
% & {\scriptsize 408, 3, 42, 2, 2, 401, 87}
% & 230
% & {\scriptsize 2, 10, 38, 22, 87} \\
% \multicolumn{2}{l}{Events (Billions)} & 160 & 30 & 20 \\
% \hline
% \end{tabular}
% \label{tab:stats}
% \end{table}

In addition to sequential event data, we use a set of pre-existing tabular user features available in the production system. For each user and month, these features are basically computed by aggregating raw behavioral signals over predefined temporal windows aligned with the given month and include both numerical and categorical variables. No additional feature construction is performed as part of this work. 

\subsubsection{Tasks and evaluation}

We evaluate the proposed approach on four downstream binary classification tasks (Task 1–Task 4), each predicting customer response to communications for a specific bank product, a key business objective for customer engagement. Performance is measured using ROC AUC on a held-out test set. All experiments are conducted on the same set of tasks and metrics, and we additionally report the average ROC AUC across all tasks for each configuration.

\subsubsection{Pretrained model configurations}

We consider multiple configurations of the pretrained sequence encoder to analyze the effect of model scale on downstream performance. Specifically, we study three model sizes that vary in depth, width, and maximum context length:
\begin{itemize}
\item \textbf{2M parameters (small)}: 6 layers, 4 attention heads, hidden size 128, context length 256
\item \textbf{8M parameters (medium)}: 8 layers, 6 attention heads, hidden size 256, context length 1024  
\item \textbf{42M parameters (large)}: 12 layers, 6 attention heads, hidden size 512, context length 2048
\end{itemize}
Unless stated otherwise, we use the medium-sized 8M configuration as the default pretrained backbone in our experiments.

\subsubsection{Pretraining setup}

We train the sequence encoder using a batch size of 32 with gradient accumulation over 16 steps, dropout rate of 0.15, and the AdamW optimizer (learning rate $10^{-3}$, weight decay $10^{-2}$, $\epsilon=10^{-4}$). We apply a cosine learning-rate schedule with 10{,}000 warmup steps and train the model for 2-10 epochs depending on the model size. Gradient clipping with a maximum norm of 1.0 is applied throughout training.

Training is distributed across 4–32 NVIDIA A100 80GB GPUs using Distributed Data Parallel (DDP). 
The best model checkpoint is selected based on validation loss.

\subsubsection{Downstream training setup}

All downstream tabular models are trained using a fixed set of hyperparameters across all experiments.
Specifically, we use a batch size of 4096,
an embedding dimension of $d_{\text{emb}}=64$, 
three transformer layers with four attention heads,
a learning rate of $0.004$, 
weight decay of $0.01$, 
6{,}000 warmup steps.
Models are trained for 40 epochs.

Training is distributed across 4 NVIDIA A100 80GB GPUs using Distributed Data Parallel (DDP). For evaluation, we select the model checkpoint with the best validation performance and report the corresponding test-set metric.

%==================================

\subsection{Comparison with a Feature-Based Baseline}

We first compare the proposed approach against a strong baseline built on hand-engineered user features that are widely used in our production systems. They have been developed and refined over an extended period of time and capture domain knowledge accumulated across multiple teams and use cases. As a result, models trained on these features present a highly competitive baseline that is difficult to outperform. Even relatively small improvements in terms of ROC AUC translate into measurable gains in business value.

Table~\ref{tab:main} reports results on four downstream prediction tasks, as well as the average performance across all tasks. We consider three variants of input representation: hand-engineered features, frozen embeddings produced by a pretrained model with 42M parameters, and a combination of pretrained embeddings and engineered features. When used in isolation, pretrained embeddings can outperform or underperform engineered features depending on the task. However, their combination is consistently better than feature-based baseline, leading to an average ROC AUC improvement of 0.009.

\begin{table}[h]
\centering
\caption{Comparison of pretrained model (42M large) with previously used approach based on engineered features.}
\begin{tabular}{lccccc}
\toprule
Approach & Task 1 & Task 2 & Task 3 & Task 4 & Avg \\
\midrule
Engineered features       & 0.681 & 0.732 & 0.891 & \textbf{0.866}  & 0.792 \\
Pretrained model  & 0.704 & 0.727 & 0.872 & 0.846  & 0.787 \\
Pretrained + features   & \textbf{0.705} & \textbf{0.742} & \textbf{0.892} & 0.864 & \textbf{0.801} \\
\bottomrule
\end{tabular}
\label{tab:main}
\end{table}

\subsection{Scaling Properties}

We further investigate the scaling properties of the pretrained model to assess whether increasing model size leads to performance improvements, as observed in other domains and prior work on foundation models~\cite{khrylchenko2025scaling,yeh2025treasure,braithwaite2025your}.

\subsubsection{Scaling model capacity}

We compare three model sizes (2M, 8M, and 42M) and evaluate each in isolation and in combination with TabNN operating on engineered features. Table~\ref{tab:scaling} reports the average ROC AUC across all downstream tasks. As expected, larger models achieve higher ROC AUC, and adding TabNN yields consistent improvements at every scale. Since engineered features already capture a substantial amount of relevant information, increasing the capacity of the pretrained model has a smaller effect in the combined setting (from 0.792 to 0.801 when scaling from 2M to 42M) compared to using pretrained embeddings alone (from 0.750 to 0.787). Notably, improvements over the feature-based baseline (0.792) become significant only at a sufficient model scale.

While further increasing model size could potentially lead to additional gains, practical constraints in the production environment currently limit the feasibility of deploying larger models. We therefore choose the 42M configuration as a reasonable trade-off between performance and operational cost.

\begin{table}[h]
\centering
\caption{The effect of the pretrained model size on the average ROC AUC metric with and without TabNN operating on engineered features.}
\begin{tabular}{lccc}
\toprule
Model Scale & Pretrained only & Pretrained + TabNN (features) \\
\midrule
2M small   & 0.750 & 0.792 \\
8M medium  & 0.774 & 0.796 \\
42M large  & \textbf{0.787} & \textbf{0.801} \\
\bottomrule
\end{tabular}
\label{tab:scaling}
\end{table}

\subsubsection{Effect of context length scaling}

We additionally study the effect of scaling only context length using the medium-sized (8M) model, varying the maximum sequence length from 256 to 1024 events. As shown in Table~\ref{tab:context}, longer user histories consistently improve ROC AUC across all four tasks, indicating that the model benefits from access to extended temporal context.

\begin{table}[h]
\centering
\caption{The effect of the transformer context length on the ROC AUC metric for various tasks. The results are shown for the 8M medium model.}
\begin{tabular}{rccccc}
\toprule
Context length& Task 1 & Task 2 & Task 3 & Task 4 & Avg. \\
\midrule
256  & 0.696 & 0.717 & 0.886 & 0.862  & 0.790 \\
512  & 0.698 & 0.725 & 0.886 & 0.859  & 0.792 \\
1024 & \textbf{0.702} & \textbf{0.727} & \textbf{0.887} & \textbf{0.864} & \textbf{0.795} \\
\bottomrule
\end{tabular}
\label{tab:context}
\end{table}

\subsection{Impact of pretraining}

To analyze the effect of self-supervised pretraining, we compare our approach against the same model architecture trained end-to-end in a supervised manner to directly predict downstream labels from user histories and engineered features. Table~\ref{tab:pretraining} shows that self-supervised pretraining consistently outperforms supervised training across all four downstream tasks, with an average ROC AUC improvement of 0.004. The largest gain is observed on Task 1, while Tasks 2–4 exhibit smaller but stable improvements. This behavior is expected in the considered data regime, where the amount of labeled data available for downstream tasks is substantially smaller than the unlabeled data used for pretraining.

\begin{table}[h]
\centering
\caption{Comparison of the ROC AUC metric for self-supervised pretraining and supervised learning on the downstream tasks with the same architecture from scratch. The results are shown for the 8M medium model.}
\begin{tabular}{lccccc}
\toprule
Method & Task 1 & Task 2 & Task 3 & Task 4 & Avg \\
\midrule
Pretrained & \textbf{0.693} & \textbf{0.736} & \textbf{0.891} & \textbf{0.859} & \textbf{0.795} \\
Supervised & 0.684 & 0.733 & 0.886 & 0.858 & 0.791 \\
\bottomrule
\end{tabular}
\label{tab:pretraining}
\end{table}

As a result, a single pretrained backbone can be reused across multiple downstream tasks in a frozen form, reducing model development and maintenance costs while simultaneously improving predictive quality.

\subsection{Contribution of Multiple Behavioral Modalities}

We analyze the contribution of different input data sources, using a model trained on Transactions (Txn-only) as a baseline, since it is the primary and most informative data source for the considered downstream tasks. Table~\ref{tab:sources} compares three configurations: a Txn-only model, a model combining Transactions and Communications (Txn + Comm), and a model that additionally incorporates Clickstream (Txn + Comm + Click), which corresponds to our final multimodal setup.

As expected, the impact of additional modalities varies across tasks, reflecting differences in the relevance of behavioral signals for each prediction objective. Nevertheless, multimodal models match or outperform the Txn-only baseline across all tasks and achieve the strongest gains on Task 2, where the Txn + Comm + Click configuration outperforms the Txn-only model by more than 0.01 ROC AUC. These results indicate that communication and clickstream data provide complementary signals beyond transactional data.

\begin{table}[h]
\centering
\caption{The impact of combining different data sources on the ROC AUC metric. The results are shown for the 8M medium model.}
\begin{tabular}{lccccc}
\toprule
Data sources & Task 1 & Task 2 & Task 3 & Task 4 & Avg \\
\midrule
Txn & \textbf{0.690} & 0.729 & 0.886 & 0.861 & 0.792 \\
Txn + Comm & 0.688 & 0.732 & 0.889 & 0.862 & 0.793 \\
Txn + Comm + Click & \textbf{0.690} & \textbf{0.741} & \textbf{0.890} & \textbf{0.862} & \textbf{0.796} \\
\bottomrule
\end{tabular}
\label{tab:sources}
\end{table}

\subsection{Comparison of Fusion Strategies}

We compare the early fusion strategy adopted in our main experiments with two late fusion approaches for combining heterogeneous sequential modalities.

In the first late fusion variant (Late fusion with hiddens), we pretrain separate sequential encoders for each modality, extract their hidden representations, and concatenate them before passing the resulting representation to the downstream model together with standard tabular features. In the second variant (Late fusion with scores), we again pretrain separate sequential encoders, but pair each of them with its own downstream model. The resulting prediction scores are then combined using a logistic regression layer trained on top of the individual model outputs.

In contrast, the early fusion model uses a single multimodal sequential encoder that jointly processes all modalities and produces a shared representation, which is subsequently consumed by a downstream model.

Table~\ref{tab:fusion} shows that early fusion achieves the highest ROC AUC on three of the four tasks (Task1, Task3, Task4), with noticeable gains on Task 1 and Task 4 and comparable performance to the best late fusion variant on Task2 and Task 3.
On average across all four tasks, early fusion delivers the best overall performance.

\begin{table}[h]
\centering
\caption{Comparison of the ROC AUC metric for different fusion strategies. The results are shown for the 8M medium model.}
\setlength{\tabcolsep}{4pt}
\begin{tabular}{lccccc}
\toprule
Method & Task 1 & Task 2 & Task 3 & Task 4 & Avg \\
\midrule
Late fusion (hiddens) & 0.687 & \textbf{0.740} & 0.892 & 0.866 & 0.796 \\
Late fusion (scores) & 0.689 & 0.731 & 0.889 & 0.859 & 0.792 \\
Early fusion & \textbf{0.691} & 0.739 & \textbf{0.892} & \textbf{0.868} & \textbf{0.797} \\
\bottomrule
\end{tabular}
\label{tab:fusion}
\end{table}

%% file: sections/7_online_results.tex
We evaluated the proposed approach in A/B tests on core banking products, where communications were delivered through three channels: push notifications, SMS, and telemarketing. As the target online metric, we used the Net Present Value (NPV), and we report the relative improvement of the treatment group over the control group:
\[
\Delta = \frac{\mathrm{NPV}_{\mathrm{treat}} - \mathrm{NPV}_{\mathrm{ctrl}}}{\mathrm{NPV}_{\mathrm{ctrl}}}.
\]

Model scores produced by our approach were consumed by an existing production optimizer. For each client–product–channel candidate, a proxy for expected incremental profit was computed as $\text{score} \times \text{NPV}$. The optimizer then built a communication plan that maximizes total NPV subject to business constraints (e.g., channel capacity limits, contact policies). 

For online evaluation, we deployed the proposed model for 6 core products (out of more than 40 products offered by the company). These products were selected based on the highest historical NPV, as they contribute the most to total revenue and provide sufficient traffic for statistically reliable A/B evaluation. We report the impact on total NPV across all products because the optimizer jointly allocates limited contact capacity across competing offers, so changes in scores for a subset of products can shift exposure and revenue among the remaining products.
As the control group, we used the production Gradient Boosting Decision Trees (GBDT) models deployed at the time of the experiment. The treatment group used the proposed architecture, while keeping all remaining business rules and operational constraints unchanged. Clients were randomly assigned to control and treatment groups, with approximately 10\% of the total client base in the treatment group and the remaining population in the control group. Statistical significance was assessed using a t-test with a significance level of 0.05, and the experiment duration (2 months) was chosen to ensure sufficient power given the variability of NPV.

Across products and communication channels, the proposed architecture achieved an average \textbf{$+1\%$ uplift} in total NPV relative to the GBDT baseline. While the absolute uplift value may appear modest, it corresponds to a substantial absolute revenue gain at the system scale. Following successful A/B testing, we transitioned the production scoring pipelines for the evaluated products to the proposed approach, which has since been adopted as the default scoring model.

%% file: sections/8_conclusion.tex
In this work, we introduced a foundation model for multimodal behavioral data that jointly leverages transactional histories, clickstream signals, and communication events. The proposed approach is based on pretraining a transformer-based sequence encoder on unified event sequences and reusing it as a frozen shared representation backbone across diverse downstream tasks. Extensive offline experiments demonstrate the effectiveness of the proposed approach. The proposed system has been successfully deployed at scale in production and validated through online A/B testing, resulting in measurable improvements in business metrics and confirming the practical value of foundation models for large-scale financial applications.

%% file: sections/bibliography.bib
@inproceedings{inproceedings_dcn,
author = {Wang, Ruoxi and Shivanna, Rakesh and Cheng, Derek and Jain, Sagar and Lin, Dong and Hong, Lichan and Chi, Ed},
year = {2021},
month = {04},
pages = {1785-1797},
title = {DCN V2: Improved Deep \& Cross Network and Practical Lessons for Web-scale Learning to Rank Systems},
doi = {10.1145/3442381.3450078}
}

@inproceedings{
gorishniy2025tabm,
title={TabM: Advancing Tabular Deep Learning with Parameter-Efficient Ensembling},
author={Yury Gorishniy and Akim Kotelnikov and Artem Babenko},
booktitle={The Thirteenth International Conference on Learning Representations},
year={2025},
url={https://openreview.net/forum?id=Sd4wYYOhmY}
}

@misc{yan2024makingpretrainedlanguagemodels,
      title={Making Pre-trained Language Models Great on Tabular Prediction}, 
      author={Jiahuan Yan and Bo Zheng and Hongxia Xu and Yiheng Zhu and Danny Z. Chen and Jimeng Sun and Jian Wu and Jintai Chen},
      year={2024},
      eprint={2403.01841},
      archivePrefix={arXiv},
      primaryClass={cs.CL},
      url={https://arxiv.org/abs/2403.01841}, 
}

@article{radford2019language,
  author = {Radford, Alec and Wu, Jeffrey and Child, Rewon and Luan, David and Amodei, Dario and Sutskever, Ilya},
  journal = {OpenAI},
  note = {Accessed: 2024-11-15},
  title = {Language Models are Unsupervised Multitask Learners},
  url = {https://cdn.openai.com/better-language-models/language_models_are_unsupervised_multitask_learners.pdf},
  year = 2019
}

@misc{klambauer2017selfnormalizingneuralnetworks,
      title={Self-Normalizing Neural Networks}, 
      author={Günter Klambauer and Thomas Unterthiner and Andreas Mayr and Sepp Hochreiter},
      year={2017},
      eprint={1706.02515},
      archivePrefix={arXiv},
      primaryClass={cs.LG},
      url={https://arxiv.org/abs/1706.02515}, 
}

@inproceedings{pancha2022pinnerformer,
  title={PinnerFormer: Sequence Modeling for User Representation at Pinterest},
  author={Pancha, Nikil and Zhai, Andrew and Leskovec, Jure and Rosenberg, Charles},
  booktitle={Proceedings of the 28th ACM SIGKDD conference on knowledge discovery and data mining},
  pages={3702--3712},
  year={2022}
}

@inproceedings{xia2023transact,
  title={TransAct: Transformer-based Realtime User Action Model for Recommendation at Pinterest},
  author={Xia, Xue and Eksombatchai, Pong and Pancha, Nikil and Badani, Dhruvil Deven and Wang, Po-Wei and Gu, Neng and Joshi, Saurabh Vishwas and Farahpour, Nazanin and Zhang, Zhiyuan and Zhai, Andrew},
  booktitle={Proceedings of the 29th ACM SIGKDD Conference on Knowledge Discovery and Data Mining},
  pages={5249--5259},
  year={2023}
}

@inproceedings{chen2025pinfm,
  title={PinFM: Foundation Model for User Activity Sequences at a Billion-scale Visual Discovery Platform},
  author={Chen, Xiangyi and Rajesh, Kousik and Lawhon, Matthew and Wang, Zelun and Li, Hanyu and Li, Haomiao and Joshi, Saurabh Vishwas and Eksombatchai, Pong and Yang, Jaewon and Hsu, Yi-Ping and others},
  booktitle={Proceedings of the Nineteenth ACM Conference on Recommender Systems},
  pages={381--390},
  year={2025}
}

@article{khrylchenko2025scaling,
  title={Scaling Recommender Transformers to One Billion Parameters},
  author={Khrylchenko, Kirill and Matveev, Artem and Makeev, Sergei and Baikalov, Vladimir},
  journal={arXiv preprint arXiv:2507.15994},
  year={2025}
}

@inproceedings{lyu2025dv365,
  title={DV365: Extremely Long User History Modeling at Instagram},
  author={Lyu, Wenhan and Tyagi, Devashish and Yang, Yihang and Li, Ziwei and Somani, Ajay and Shanmugasundaram, Karthikeyan and Andrejevic, Nikola and Adeputra, Ferdi and Zeng, Curtis and Singh, Arun K and others},
  booktitle={Proceedings of the 31st ACM SIGKDD Conference on Knowledge Discovery and Data Mining V. 2},
  pages={4717--4727},
  year={2025}
}

@inproceedings{dabrowski2025recsys,
  title={RecSys Challenge 2025: Universal Behavioral Profiles for Recommender Systems},
  author={Dabrowski, Jacek and Janicka, Maria and Sienkiewicz, Lukasz and Stomfai, Gergely and Jannach, Dietmar and Barile, Francesco and Polignano, Marco and Pomo, Claudio and Srivastava, Abhishek},
  booktitle={Proceedings of the Nineteenth ACM Conference on Recommender Systems},
  pages={1389--1393},
  year={2025}
}

@incollection{sawada2025toward,
  title={Toward Universal User Representations: Contrastive Learning with Transformers and Embedding Ensembles},
  author={Sawada, Yuki and Hasegawa, Rintaro and Nagatsuma, Yuhi and Takei, Shugo and Yonekawa, Kazuhito and Auchi, Hiromu},
  booktitle={Proceedings of the Recommender Systems Challenge 2025},
  pages={51--55},
  year={2025}
}

@incollection{makeev2025blending,
  title={Blending Sequential Embeddings, Graphs, and Engineered Features: 4th Place Solution in RecSys Challenge 2025},
  author={Makeev, Sergei and Andreev, Alexandr and Baikalov, Vladimir and Tytskiy, Vladislav and Krasilnikov, Aleksei and Khrylchenko, Kirill},
  booktitle={Proceedings of the Recommender Systems Challenge 2025},
  pages={21--25},
  year={2025}
}

@incollection{klenitskiy2025encode,
  title={Encode Me If You Can: Learning Universal User Representations via Event Sequence Autoencoding},
  author={Klenitskiy, Anton and Fatkulin, Artem and Denisova, Daria and Pembek, Anton and Vasilev, Alexey},
  booktitle={Proceedings of the Recommender Systems Challenge 2025},
  pages={26--30},
  year={2025}
}

@inproceedings{babaev2022coles,
  title={CoLES: Contrastive Learning for Event Sequences with Self-Supervision},
  author={Babaev, Dmitrii and Ovsov, Nikita and Kireev, Ivan and Ivanova, Maria and Gusev, Gleb and Nazarov, Ivan and Tuzhilin, Alexander},
  booktitle={Proceedings of the 2022 International Conference on Management of Data},
  pages={1190--1199},
  year={2022}
}

@inproceedings{skalski2023towards,
  title={Towards a Foundation Purchasing Model: Pretrained Generative Autoregression on Transaction Sequences},
  author={Skalski, Piotr and Sutton, David and Burrell, Stuart and Perez, Iker and Wong, Jason},
  booktitle={Proceedings of the Fourth ACM International Conference on AI in Finance},
  pages={141--149},
  year={2023}
}

@article{yeh2025treasure,
  title={TREASURE: A Transformer-Based Foundation Model for High-Volume Transaction Understanding},
  author={Yeh, Chin-Chia Michael and Saini, Uday Singh and Dai, Xin and Fan, Xiran and Jain, Shubham and Fan, Yujie and Sun, Jiarui and Wang, Junpeng and Pan, Menghai and Dou, Yingtong and others},
  journal={arXiv preprint arXiv:2511.19693},
  year={2025}
}

@article{braithwaite2025your,
  title={Your Spending Needs Attention: Modeling Financial Habits with Transformers},
  author={Braithwaite, DT and Cavalcanti, Misael and McEver, R Austin and Udagawa, Hiroto and Silva, Daniel and Ramanath, Rohan and Meneses, Felipe and Yoshida, Arissa and Wingert, Evan and Ramos, Matheus and others},
  journal={arXiv preprint arXiv:2507.23267},
  year={2025}
}

@unknown{ft_transformer,
author = {Dai, Huangliang and Wu, Shixun and Zhao, Hairui and Huang, Jiajun and Jian, Zizhe and Zhu, Yue and Hu, Haiyang},
year = {2025},
month = {04},
pages = {},
title = {FT-Transformer: Resilient and Reliable Transformer with End-to-End Fault Tolerant Attention},
doi = {10.48550/arXiv.2504.02211}
}

@article{karpukhin2024detecting,
  title={Detecting the Future: All-at-Once Event Sequence Forecasting with Horizon Matching},
  author={Karpukhin, Ivan and Savchenko, Andrey},
  journal={arXiv preprint arXiv:2408.13131},
  year={2024}
}

@inproceedings{mollaev2025multimodal,
  title={Multimodal Banking Dataset: Understanding Client Needs through Event Sequences},
  author={Mollaev, Dzhambulat and Kireev, Ivan and Orlov, Mikhail and Kostin, Alexander and Karpukhin, Ivan and Postnova, Maria and Gusev, Gleb and Savchenko, Andrey},
  booktitle={Proceedings of the 34th ACM International Conference on Information and Knowledge Management},
  pages={6476--6480},
  year={2025}
}

@article{karpukhin2025ht,
  title={HT-Transformer: Event Sequences Classification by Accumulating Prefix Information with History Tokens},
  author={Karpukhin, Ivan and Savchenko, Andrey},
  journal={arXiv preprint arXiv:2508.01474},
  year={2025}
}

@article{ostroukhov2026pragma,
  title={PRAGMA: Revolut Foundation Model},
  author={Ostroukhov, Maxim and Mikhailov, Ruslan and Iashin, Vladimir and Sokolov, Artem and Akshonov, Andrei and Protasov, Vitaly and Beloborodov, Dmitrii and Mullin, Vince and Enzmann, Roman Yokunda and Kolovos, Georgios and others},
  journal={arXiv preprint arXiv:2604.08649},
  year={2026}
}

@article{dou2025transactiongpt,
  title={TransactionGPT},
  author={Dou, Yingtong and Jiang, Zhimeng and Zhang, Tianyi and Hu, Mingzhi and Xu, Zhichao and Jain, Shubham and Saini, Uday Singh and Fan, Xiran and Sun, Jiarui and Pan, Menghai and others},
  journal={arXiv preprint arXiv:2511.08939},
  year={2025}
}

@article{jha2012employing,
  title={Employing transaction aggregation strategy to detect credit card fraud},
  author={Jha, Sanjeev and Guillen, Montserrat and Westland, J Christopher},
  journal={Expert systems with applications},
  volume={39},
  number={16},
  pages={12650--12657},
  year={2012},
  publisher={Elsevier}
}

@article{polleti2025open,
  title={Open Banking Foundational Model: Learning Language Representations from Few Financial Transactions},
  author={Polleti, Gustavo and Santana, Marlesson and Fontes, Eduardo},
  journal={arXiv preprint arXiv:2511.12154},
  year={2025}
}

@inproceedings{guo2025efficient,
  title={Efficient Multi-Expert Tabular Language Model for Banking},
  author={Guo, Yue and Zhang, Wentao and Zhang, Xiaojun and Zheng, Vincent W and Yang, Yi},
  booktitle={Proceedings of the 31st ACM SIGKDD Conference on Knowledge Discovery and Data Mining V. 1},
  pages={2271--2281},
  year={2025}
}

@article{raman2024scalable,
  title={Scalable Representation Learning for Multimodal Tabular Transactions},
  author={Raman, Natraj and Ganesh, Sumitra and Veloso, Manuela},
  journal={arXiv preprint arXiv:2410.07851},
  year={2024}
}

@inproceedings{wang2023sequence,
  title={Sequence As Genes: An User Behavior Modeling Framework for Fraud Transaction Detection in E-commerce},
  author={Wang, Ziming and Wu, Qianru and Zheng, Baolin and Wang, Junjie and Huang, Kaiyu and Shi, Yanjie},
  booktitle={Proceedings of the 29th ACM SIGKDD Conference on Knowledge Discovery and Data Mining},
  pages={5194--5203},
  year={2023}
}

@article{gong2025behavegpt,
  title={BehaveGPT: A Foundation Model for Large-scale User Behavior Modeling},
  author={Gong, Jiahui and Ding, Jingtao and Meng, Fanjin and Yang, Chen and Chen, Hong and Wang, Zuojian and Lu, Haisheng and Li, Yong},
  journal={arXiv preprint arXiv:2505.17631},
  year={2025}
}

@inproceedings{liu2022user,
  title={User Behavior Pre-training for Online Fraud Detection},
  author={Liu, Can and Gao, Yuncong and Sun, Li and Feng, Jinghua and Yang, Hao and Ao, Xiang},
  booktitle={Proceedings of the 28th ACM SIGKDD Conference on Knowledge Discovery and Data Mining},
  pages={3357--3365},
  year={2022}
}

@article{bazarova2025learning,
  title={Learning Transactions Representations for Information Management in Banks: Mastering Local, Global, and External Knowledge},
  author={Bazarova, Alexandra and Kovaleva, Maria and Kuleshov, Ilya and Romanenkova, Evgenia and Stepikin, Alexander and Yugay, Aleksandr and Mollaev, Dzhambulat and Kireev, Ivan and Savchenko, Andrey and Zaytsev, Alexey},
  journal={International Journal of Information Management Data Insights},
  volume={5},
  number={1},
  pages={100323},
  year={2025},
  publisher={Elsevier}
}

@inproceedings{sakhno2025pytorch,
  title={PyTorch-Lifestream: Learning Embeddings on Discrete Event Sequences},
  author={Sakhno, Artem and Kireev, Ivan and Babaev, Dmitrii and Savchenko, Maxim and Gusev, Gleb and Savchenko, Andrey},
  booktitle={Proceedings of the Thirty-Fourth International Joint Conference on Artificial Intelligence},
  pages={11104--11108},
  year={2025}
}

@inproceedings{shestov2025llm4es,
  title={LLM4ES: Learning user embeddings from event sequences via large language models},
  author={Shestov, Aleksei and Zoloev, Omar and Makarenko, Maksim and Orlov, Mikhail and Fadeev, Egor and Kireev, Ivan and Savchenko, Andrey},
  booktitle={Proceedings of the 34th ACM International Conference on Information and Knowledge Management},
  pages={5238--5242},
  year={2025}
}

@article{moskvoretskii2024mlem,
  title={MLEM: Generative and Contrastive Learning as Distinct Modalities for Event Sequences},
  author={Moskvoretskii, Viktor and Osin, Dmitry and Shvetsov, Egor and Udovichenko, Igor and Zhelnin, Maxim and Dukhovny, Andrey and Zhimerikina, Anna and Burnaev, Evgeny},
  journal={arXiv preprint arXiv:2401.15935},
  year={2024}
}
